\documentclass{article}
\pdfoutput=1 
\usepackage{arxiv}

\usepackage{threeparttable}
\usepackage{CJKutf8}
\usepackage{amsmath}
\usepackage{amssymb}

\usepackage[T1]{fontenc}    
\usepackage{hyperref}       
\usepackage{cleveref}
\usepackage{url}            
\usepackage{booktabs}       
\usepackage{amsfonts}       
\usepackage{nicefrac}       
\usepackage{microtype}      
\usepackage{lipsum}		
\usepackage{graphicx}
\usepackage{multirow}

\usepackage{float}

\title{A Practical Chinese Dependency Parser Based on A Large-scale Dataset}

\author{
Shuai Zhang \\
Baidu \\
\texttt{zhangshuai28@baidu.com} \\
\And
Lijie Wang \\
Baidu\\
\texttt{wanglijie@baidu.com} \\
\And
Ke Sun \\
Baidu \\
\texttt{sunke@baidu.com} \\
\And
Xinyan Xiao \\
Baidu \\
\texttt{xiaoxinyan@baidu.com} \\
}

\hypersetup{
pdftitle={A template for the arxiv style},
pdfsubject={q-bio.NC, q-bio.QM},
pdfauthor={David S.~Hippocampus, Elias D.~Striatum},
pdfkeywords={First keyword, Second keyword, More},
}

\begin{document}

\maketitle

\begin{abstract}

Dependency parsing is a longstanding natural language processing task, with its outputs crucial to various downstream tasks.
Recently, neural network based (NN-based) dependency parsing has achieved significant progress and obtained the state-of-the-art results.
As we all know, NN-based approaches require massive amounts of labeled training data, which is very expensive because it requires human annotation by experts. Thus few industrial-oriented dependency parser tools are publicly available.
In this report, we present Baidu Dependency Parser (DDParser), a new Chinese dependency parser trained on a large-scale manually labeled dataset called Baidu Chinese Treebank (DuCTB). DuCTB consists of about one million annotated sentences from multiple sources including search logs, Chinese newswire, various forum discourses, and conversation programs. DDParser is extended on the graph-based biaffine parser to accommodate to the characteristics of Chinese dataset. 
We conduct experiments on two test sets: the standard test set with the same distribution as the training set and the random test set sampled from other sources, and the labeled attachment scores (LAS) of them are 92.9\% and 86.9\% respectively. DDParser achieves the state-of-the-art results, and is released at \url{https://github.com/baidu/DDParser}.

\end{abstract}

\keywords{Chinese dependency parsing \and Biaffine \and Chinese treebank \and Baidu dependency parser}

\section{Introduction}

\begin{figure}[tb]
\centering
\includegraphics[width=0.7\textwidth]{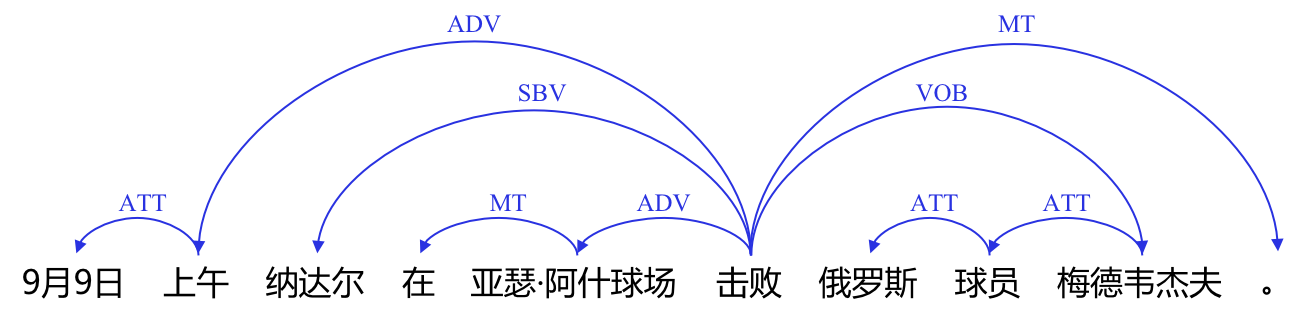}
\caption{An example of the dependency parse tree.}
\label{fig:intro_case}
\end{figure}

Dependency parsing aims to annotate sentences into a dependency tree which is designed to be easy for humans and computers alike to understand. Given an input sentence $s=w_0w_1...w_n$, a dependency tree, as depicted in Figure \ref{fig:intro_case}, is defined as $d=\{(h,m,l), 0 \leq h \leq n, 1 \leq m \leq n, l \in \pounds\}$, where $(h,m,l)$ is a dependency from the head word $w_h$ to the modifier word $w_m$ with the relation label $l \in \pounds$, and $w_0$ is a pseudo word that points to the root word. As a fundamental task in natural language processing (NLP), dependency parsing has been found to be extremely useful for a sizable number of NLP tasks, especially those involving natural language understanding in some way \cite{bowman2016fast, angeli2015leveraging, levy2014dependency, toutanova2016compositional, parikh2015grounded}.

In recent years, NN-based approaches have achieved remarkable improvement and outperformed the traditional discrete-feature based approaches in dependency parsing by a large margin \cite{chen2014fast, dyer2015transition}.
\cite{dozat2016deep} propose a simple yet effective deep biaffine graph-based parser and achieve the state-of-the-art accuracy on a variety of datasets and languages.
Based on this work, \cite{li2019self} applies the self-attention based encoder to dependency parsing as the replacement of BiLSTMs, and then make an in-depth study on the the differences between the two techniques. As we all known, labeled data is very critical for all NN-based approaches, including data size, annotation quality and so on. However, it is difficult to build a large-scale dependency parsing dataset by human annotation.

After about a decade of accumulation and innovation, Baidu has established a Chinese dependency parsing dataset (DuCTB) with a scale of nearly one million, covering multiple sources such as search logs, Chinese newswire, forum discourses. Then an effective dependency parsing tool is trained based on DuCTB, achieving the state-of-the-art results. In order to help ordinary users to obtain the syntactic and semantic information of sentences, we release our dependency parser including the source code and trained model. Our parser has three advantages: 1) the training data consists of more than 500,000 sentences\footnote{The model we released is not trained with the full training data.}, covering news, conversations and search queries, etc; 2) it outperforms other dependency parsers both on the labeled attachment score (LAS) and unlabeled attachment score (UAS); 3) it is very convenient to use, as the installation and prediction can be implemented with a single command.

We release our model and source code at \url{https://github.com/baidu/DDParser}.

\section{Dataset}
\label{sec:data}

Motivated by different syntactic theories and practices, major languages in the world often possess multiple large-scale heterogeneous treebanks. Table \ref{tab:chs_data} lists several large-scale Chinese treebanks each of which has a different annotation guideline. We introduce DuCTB  from the following aspects.
\begin{itemize}
\item \textbf{Sentence selection}. Sentences from different sources are different in the way of expression, which has certain influence on the analysis of syntactic structure. For example, the sentence from news is usually expressed in line with the syntax, but the sentence from search logs and forums are often expressed irregularly, such as inversion, ellipsis. In order to cover as many expressions as possible, we sample unlabeled sentences from as many sources as possible. The sources mainly covers two cases: 1) regular sentences, mainly from news, network reading materials; 2) irregular sentences, mainly from search logs, forum discourses, texts transformed from voice, conversation utterances. At last, we get about 1,000,000 labeled sentences. We use CONLL-X \cite{buchholz2006conll} as the data output style to represent our dataset.
\item \textbf{Annotation guideline}. The DuCTB is built for industrial applications and focuses on analyzing the syntactic structure  of the sentence other than its semantics. Our annotation guideline aims to be understood by ordinary users. Table \ref{tab:label_list} shows all labels defined in the guideline, including the definitions and the corresponding examples. Different from other treebanks, DuCTB focuses on analyzing relations between notional words, such as nouns, verbs. The empty word such as punctuation words, conjunction words, preposition words, has a relation of ``MT'' with its head. Figure \ref{fig:intro_case} shows an example of DDParser.

\end{itemize}

\begin{table*}[tb]
\caption{Large-scale Chinese treebanks.}
\renewcommand\tabcolsep{2.5pt}
\centering
\begin{tabular}{l | c|c|c|c}
\toprule
Treebanks  & Tokens & Sentences & Grammar & Sources \\
\hline
Sinica \cite{chen2003sinica} & 0.36M & 0.06M & Case grammar & Literature, life, society, art, etc. \\
\hline
CTB \cite{xue2005penn} & 1.62M & 0.07M & Phrase structure & newswire \\
\hline
TCT \cite{qiang2004annotation} & 1.00M & 0.05M &  Phrase structure & journalistic, literary, academic, etc.\\
\hline
\multirow{2}{*}{PCT \cite{zhan2012application}} & \multirow{2}{*}{1.31M} & \multirow{2}{*}{0.06M} & Phrase structure & chinese textbook, \\  &  &  &  & government documents, newswire etc. \\
\hline
HIT-CDT \cite{che2012chinese} & 0.90M & 0.05M  & Dependency structure & newswire \\
\hline
PKU-CDT \cite{qiu2014multi} & 0.33M & 0.01M & Dependency structure & newswire \\
\hline
\multirow{2}{*}{CODT \cite{guo2019construction}} & \multirow{2}{*}{-} & \multirow{2}{*}{0.13M} & \multirow{2}{*}{Dependency structure} &  Chinese textbook, comment, \\
 &  &  &  & newswire, conversation programs, etc \\
\hline
\multirow{2}{*}{DuCTB} & \multirow{2}{*}{9.95M} & \multirow{2}{*}{0.95M} & \multirow{2}{*}{Dependency structure} & search logs, forum discourses, \\
 &  &  &  & newswire, conversation programs, etc \\
\bottomrule
\end{tabular}
\label{tab:chs_data}
\end{table*}

\begin{CJK*}{UTF8}{gbsn}
\begin{table*}[tb]
\caption{Dependency relation tags in DuCTB}
\renewcommand\tabcolsep{2.5pt}
\centering
\begin{tabular}{  l | c |c} 
\hline 
Relation & Description  & Example\\
\hline  
\multirow{2}{*}{SBV} & 主语与谓语间的关系 & 他送了一本书。 (送, 他, SBV)\\
 & subject and predict & He gives a book. (gives, he, SBV)\\
\hline
\multirow{2}{*}{VOB} & 宾语与谓词间的关系 & 他送了一本书。 (送, 书, VOB)\\
 & object and predict & He gives a book. (gives, book, VOB)\\
\hline
\multirow{2}{*}{POB} & 介词与宾语间的关系 & 我把书卖了。 (把, 书, POB)\\
& preposition and object & I sold the book to her. (to, her, POB)\\
\hline
\multirow{2}*{}{ADV} & 状语与中心词间的关系 & 我今天买书了。 (买, 昨天, ADV)\\
& adverbial modifier and head word & I bought a book today. (bought, today, ADV)\\
\hline 
\multirow{2}{*}{CMP} & 补语与中心词间的关系 & 我吃多了。 (吃, 多, CMP)\\
& complement and head word & I ate too much. (ate, too much, CMP)\\
\hline  
\multirow{2}{*}{ATT} & 定语与中心词间的关系 & 他送了一本书。 (书, 一本, ATT)\\
& attribute and head word & He sent a book. (book, a, ATT)\\
\hline
\multirow{2}{*}{F} & 方位词与中心词的关系 & 在公园里玩耍。 (公园, 里, F)\\
& directional word and head word & Play in the park. (park, in, F)\\
\hline
\multirow{2}{*}{COO} & 同类型词语间关系 & 叔叔阿姨 (叔, 阿姨, COO)\\
& two coordinate words & Uncle aunt (Uncle, aunt, COO)\\
\hline
\multirow{2}{*}{DBL} & 主谓短语做宾语 & 我们邀请他玩。 (请, 他, DBL)\\
& predict and subject-predict based object & we invited him to play. (invited, him, DBL)\\
\hline
\multirow{2}{*}{DOB} & 双宾语结构 & 他送我一本书。 (送, 我, DOB)\\
& double objects & He gave me a book. (gave, me, DOB)\\
\hline
\multirow{2}{*}{VV} & 同主语的多个谓词间关系 & 他外出打篮球。 (外出, 打, VV)\\
& multiple predicts & He went to play basketball. (went, play, VV)\\
\hline
\multirow{2}{*}{IC} & 两个结构独立或关联的单句 & 你好，书店怎么走？(走, 你好, IC)\\
& two independent structure & Hello, how can I get to ...? (get, hello, IC)\\
\hline
\multirow{2}{*}{MT} & 虚词与中心词间的关系 & 他送了一本书。 (送, 了, MT)\\
& empty word and its head word & \\
\hline
\multirow{2}{*}{HED} & 指整个句子的核心 & 他送了一本书。 (ROOT, 送, HED)\\
& sentence head and pseudo word & He sent a book. (ROOT, sent, HED)\\
\hline 
\end{tabular}
\label{tab:label_list}
\end{table*}
\end{CJK*}
\section{Methods}
\label{sec:method}
We extend the biaffine parser \cite{dozat2016deep} which is the most popular method in dependency parsing task to accommodate to DuCTB dataset. At present, the biaffine parser reports the state-of-the-art results both in accuracy and inference speed, and has been used in many other models \cite{zhang2020efficient} or projects, such as LTP\footnote{\url{http://www.ltp-cloud.com}} and FastNLP\footnote{\url{https://github.com/fastnlp/fastNLP}}. We only provide high-level model descriptions for biaffine parser and refer to the source paper for details. 

\subsection{Model Architecture}
\label{ssec:model_arc}
\begin{figure}[tb]
\centering
\includegraphics[width=0.8\textwidth]{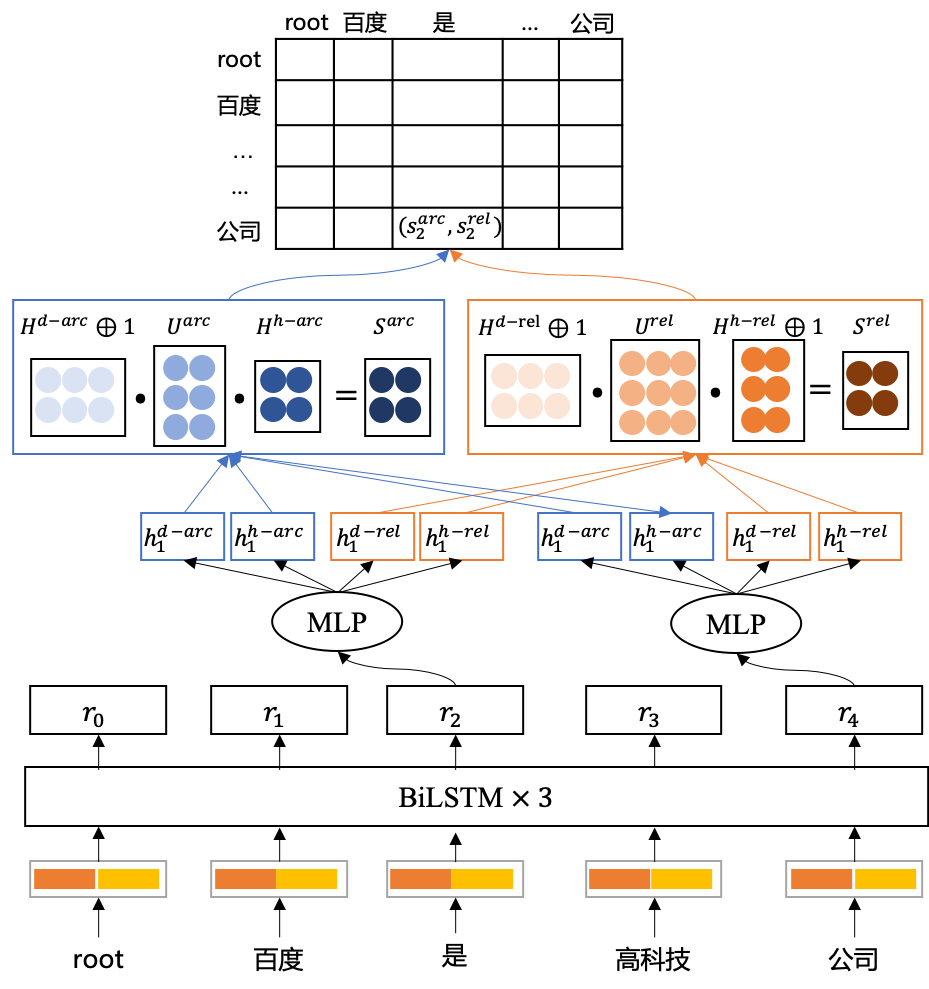}
\caption{An overview of DDParser.}
\label{fig:method_model}
\end{figure}

This subsection introduces the network architecture of our parser, as shown in Figure \ref{fig:method_model}. We will introduce its main components in detail.

\textbf{Inputs}. For the $i$th word, its input vector $e_i$ is the concatenation of the word embedding and character-level representation:
\begin{equation}
e_i = e^{word}_i \oplus CharLSTM(w_i)
\end{equation}
Where CharLSTM($w_i$) is the output vectors after feeding the character sequence into a BiLSTM layer \cite{lample2016neural}. The experimental results on DuCTB dataset show that replacing POS tag embeddings with CharLSTM($w_i$) leads to the improvement.

\textbf{BiLSTM encoder}. We employ three BiLSTM layers over the input vectors for context encoding. We denote as $r_i$ the output vector of the top-layer BiLSTM for $w_i$.

\textbf{Biaffine parser}. We apply the dependency parser of \cite{dozat2016deep} and follow most of its parameter settings. We apply dimension-reducing MLPs to each recurrent output vector $r_i$ before applying the biaffine transformation. As described in \cite{dozat2016deep}, applying smaller MLPs to the recurrent output states before the biaffine classifier has the advantage of stripping away information not relevant to the current decision. Then we use biaffine attention both in dependency arc classifier and relation classifier.  The computations of all symbols in Figure \ref{fig:method_model} are shown below.
\begin{equation}
h^{d-arc}_i = MLP^{d-arc}(r_i)
\end{equation}
\begin{equation}
h^{h-arc}_i = MLP^{h-arc}(r_i)
\end{equation}
\begin{equation}
h^{d-rel}_i = MLP^{d-rel}(r_i)
\end{equation}
\begin{equation}
h^{h-rel}_i = MLP^{h-rel}(r_i)
\end{equation}
\begin{equation}
S^{arc} = (H^{d-arc} \oplus I)U^{arc}H^{h-arc}
\end{equation}
\begin{equation}
S^{rel} = (H^{d-rel} \oplus I)U^{rel}((H^{h-rel})^T \oplus I)^T
\end{equation}

\textbf{Decoder}. We use the first-order Eisner algorithm \cite{eisner2000bilexical} in the decoder to ensure that the output is a projection tree. According to our analysis on the outputs, we find that the outputs of most sentences are projective trees. Thus we propose a strategy to judge whether the output is a legal projection tree before using the Eisner algorithm. Based on the dependency tree built by biaffine parser, we get a word sequence through the in-order traversal of the tree. The output is a projection tree only if the word sequence is in order.

\subsection{Our implementation}
\label{ssec:our_imp}

Our parser is implemented in PaddlePaddle\footnote{\url{https://www.paddlepaddle.org.cn}}. Model parameter settings are shown in Table \ref{tab:para_set}.

\begin{table*}[!htb]
\caption{Model parameters.}
\renewcommand\tabcolsep{2.5pt}
\centering
\begin{tabular}{lc|lc}
\toprule
Parameter & Value & Parameter & Value \\
\hline
Word embedding size & 300 & Word dropout & 0.33  \\
Char embedding size & 50 & Char dropout & 0.33 \\
LSTM size & 400 & LSTM dropout & 0.33 \\
Arc MLP size & 500 & Arc MLP dropout & 0.33 \\
Relation MLP size & 100 & Relation MLP dropout & 0.33 \\
LSTM depth & 3 & MLP depth & 1 \\
Optimization & Adam & Learning rate & 2e-3 \\
\bottomrule
\end{tabular}
\label{tab:para_set}
\end{table*}
\section{Experimental}
\subsection{Evaluations}
On all datasets, we use the standard labeled attachment scores (LAS) and unlabeled attachment scores (UAS) to measure the parsing accuracy. Both LAS and UAS are standard evaluation metric in dependency parsing tasks. LAS is the percentage of words that get both the correct syntactic head and dependency relation, and UAS is the percentage of words that get the correct syntactic head.

\begin{equation}
    LAS = \frac{the\ number\ of\ words\ assigned \ correct\ head\ and\ relation}{total\ words}
\end{equation}
\begin{equation}
    UAS = \frac{the\ number\ of\ words\ assigned\ correct \ head}{total\ words}
\end{equation}


\subsection{Datasets}
We conduct our experiments on the Chinese Treebank 5\footnote{An extension of CTB, please refer to \url{https://catalog.ldc.upenn.edu/LDC2005T01} for details} (CTB5) and Baidu Chinese Treebank (DuCTB).
\begin{itemize}
\item \textbf{CTB5} consists of 18,786 sentences which are split into 16,074/803/1,905 for train/dev/test sets. Its content comes from the newswire sources including Xinhua, Information Services Department of HKSAR and Taiwan Sinorama magazine.
\item \textbf{DuCTB} consists of about one million sentences from multiple sources, such as search logs, Chinese newswire, various forum discourses, conversation programs. The standard test set with the same distribution with the training set consists of 2,592 sentences.
\end{itemize}

In order to test the model's ability of generalization to sentences from new sources, we build a random test set whose sentences are randomly sampled from other sources which are not covered by training data. This new test set includes 500 sentences.

%



\subsection{Results}
For the CTB5 dataset, we use gold POS tags. We represent each word using word embedding and POS embedding, the dimension of each is 300 and 100 respectively. For the DuCTB dataset, we represent each word using word embedding and char embedding \cite{karpathy2015unreasonable}, the dimension of each is 300 and 50 respectively. Other parameter settings refer to the implementation in Section $\$$\ref{ssec:our_imp}. We give the performances of our parser on two datasets in the Table \ref{tab:uas_las_ctb_ductb}.


\begin{table}[H]
\renewcommand\tabcolsep{2.5pt}
\caption{Test accuracy on CTB5 and DuCTB.}
\centering
\begin{tabular}{ c | c | c } 
\hline 
Test Set & UAS & LAS\\
\hline
CTB5 test& 90.3\% & 89.1\% \\
DuCTB standard test & 94.8\% & 92.9\% \\
DuCTB random test & 89.7\% & 86.9\% \\
\hline
\end{tabular}
\label{tab:uas_las_ctb_ductb}
\end{table}

Meanwhile, we give a comparison with several publicly available parser tools in Table \ref{tab:random_test}, where all tools are evaluated on the random test set according to their own annotation guidelines. We can see that DDparser outperforms other tools.
\begin{table}[H]
\renewcommand\tabcolsep{2.5pt}
\caption{Test accuracy on CTB5 and DuCTB.}
\centering
\begin{tabular}{ c | c | c } 
\hline 
Model & UAS & LAS\\
\hline
DDParser & 89.7\% & 86.9\% \\
parser1 & 88.8\% & 86.5\% \\
parser2 & 78.6\% & 75.2\% \\
\hline
\end{tabular}

\label{tab:random_test}
\end{table}

\bibliographystyle{uns/rt}

\bibliographystyle{IEEEtran}

%
%
%
%

\end{document}